\documentclass[sn-mathphys]{sn-jnl}

\jyear{2023}

\usepackage[export]{adjustbox}
\usepackage{amsfonts}
\usepackage{amsmath}
\usepackage{amssymb}
\usepackage[greek,english]{babel}
\usepackage{blkarray}
\usepackage{hyperref}
\usepackage[utf8]{inputenc} 
\usepackage{setspace}
\usepackage{wrapfig}
\usepackage{xspace}

\usepackage{mathtools}
\DeclarePairedDelimiter\bra{\langle}{\rvert}
\DeclarePairedDelimiter\ket{\lvert}{\rangle}
\DeclarePairedDelimiterX\braket[2]{\langle}{\rangle}{#1 \delimsize\vert #2}

\newcommand{\ov}{\overline}
\renewcommand{\citet}{\cite}

\newcommand{\tr}{\mathrm{tr}}
\newcommand{\ot}{\otimes}

\providecommand{\pgfsyspdfmark}[3]{}

\begin{document}

\title[Near-Term Advances in Quantum Natural Language Processing]{Near-Term Advances in Quantum Natural Language Processing}

\author*{\fnm{Dominic} \sur{Widdows}}\email{widdows@ionq.com}
\author{\fnm{Aaranya} \sur{Alexander}}
\author{\fnm{Daiwei} \sur{Zhu}}
\author{\fnm{Chase} \sur{Zimmerman}}
\author{\fnm{Arunava} \sur{Majumder}}

\affil{\orgname{IonQ Inc.}, \city{College Park}, \state{Maryland}, \country{USA} \linebreak \small{Preprint from 2023. \linebreak See \url{https://link.springer.com/article/10.1007/s10472-024-09940-y} for updated official version}}


\abstract{This paper describes experiments showing that some tasks in natural language
processing (NLP) can already be performed using quantum computers, though so far only with small datasets.

We demonstrate various approaches to topic classification. 
The first uses an explicit word-based approach, in which word-topic  weights are
implemented as fractional rotations of individual qubits, and a phrase is classified
based on the accumulation of these weights onto a scoring qubit using entangling quantum gates.
This is compared with more scalable quantum encodings of word embedding vectors, which are used
to compute kernel values in a quantum support vector machine: this approach achieved
an average of 62\% accuracy on classification tasks involving over 10000 words, which is 
the largest such quantum computing experiment to date.

We describe a quantum probability approach to bigram modeling that can be applied to understand sequences of words and formal concepts, 
investigate a generative approximation to these distributions using a quantum circuit Born machine, and introduce
an approach to ambiguity resolution in verb-noun composition using single-qubit rotations for simple nouns and
2-qubit entangling gates for simple verbs.

The smaller systems presented have been run successfully on physical 
quantum computers, and the larger ones have been simulated. 
We show that statistically meaningful results can be obtained,
but are much more difficult to predict using real datasets than 
using artificial language examples from previous quantum NLP research. 

Related NLP research is compared, partly with respect to contemporary 
challenges including informal language, fluency, and truthfulness.}


\maketitle

\section{Introduction}

Natural Language Processing (NLP) is a promising area of application using quantum computers. 
This paper reports on progress in developing and running 
NLP experiments on trapped-ion quantum computers mainly during the year 2022. The experimental applications include 
topic classification, bigram modeling, and ambiguity resolution.

More generally, building real artificial general intelligence is a core goal of computer science. 
Reasons for believing that quantum models and quantum computing 
may contribute to building more fully intelligent systems are varied, 
and are discussed in Section \ref{promise_sec}. 
Our approach to building early-stage language processing systems
on quantum computers is motivated by these goals. Questions this entails include: 
what properties are lacking in artificial intelligence (AI) today; whether these gaps
can be addressed with quantum approaches; 
what core mathematical / language operations can be identified; and 
for each operation, what are the simplest effective implementations?
These questions help to motivate medium-term challenges and immediate progress.

The examples presented in this paper include word-based topic classification in Section 
\ref{classification_sec}, bigram sequence modelling in Section \ref{bigram_sec},
and ambiguity resolution in verb-noun composition in Section \ref{ambiguity_sec}.
Progress in related systems and fields is surveyed in Section \ref{background_sec},
focusing partly on the question of whether natural language grammar and syntax needs
to be modeled explicitly in quantum NLP. Thanks partly to progress with more mainstream NLP language models,
we argue that factual accuracy is more central and challenging than aligning semantics with syntax explicitly.
The conclusion in Section \ref{conclusion_sec} revisits core questions in scale and accuracy:
our experiments have processed large enough real datasets to produce statistically meaningful results,
though considerably smaller than would be needed to compete with standard classical methods on natural language data.

\section{The Promise of Quantum NLP}
\label{promise_sec}

Overlaps between quantum theory and NLP are many and varied, ranging from structural and qualitative
similarities to directly sharing mathematical techniques.

\subsection{Structural Similarities}

Language and quantum mechanics have notable qualitative similarities \citep{widdows2003mathematical}. 
Most words have several possible meanings or could refer
to several different things. Even if we can predict the range of options in advance, we do not typically know which interpretation
of a word or phrase is appropriate until we encounter it in context. However, once a word is observed
in context, one of the available meanings is typically selected, and this selection tends to remain fixed until the context changes.
Similarly, quantum mechanics predicts the probabilities of different outcomes, not which outcome is actually found, unless the system is already in a fixed state that is aligned with the measurement being performed. When measured, the system is observed to be
in a fixed state, and if measured again, the same result will be given --- the outcome remains fixed until something changes it
\cite[Ch 2]{dirac1930quantum}. Even if we prefer precise deterministic theories in physics and linguistics, nature and language are
unpredictable in practice. 

Features found in quantum theory have been explored in other human sciences, noting similar patterns. 
The nature of money in economics and finance is one example: David Orrell's textbook {\it Quantum Economics and Finance} argues that
credits and debts show similarities to the concept of quantum entanglement, and that
money, price, and value are uncertain properties that only become fixed when they are agreed and recorded in 
transactions, likening the former to quantum and the latter to classical information \citet{orrell2020quantum}. Cognitive scientists have
measured relationships between available information, context, and decision-making, that violate
classical probabilistic laws but can be represented concisely and accurately as phase interference terms in 
quantum models \citep{busemeyer2012quantum}.

Such similarities do not show us how to perform particular language processing tasks, but 
they do suggest that there is a rich space to explore, which is especially inviting for problems involving
large amounts of context and uncertainty which frustrate classical or more reductionist approaches.
For some human problems, if we look to mathematics and physics for helpful techniques, 
quantum mechanics and logic can be just as intuitive a place to start as classical mechanics and logic.

\subsection{Compositional Behaviors}

Language is compositional, in the sense that the meanings of new phrases and sentences can be derived from their parts,
even when the phrase itself has not been seen before or describes something quite unrealistic. 
For example, one may have never heard the phrase ``purple polar bear'' before,
but it immediately conjures up a mental picture of a polar bear whose color is purple.

Language composition sometimes produces similar structures from dissimilar ingredients --- for example, in the context
of hotels, ``book a room'' and ``make a reservation'' have overlapping and sometimes identical meanings, even though 
neither of the pairs ({\it book}, {\it make}) or ({\it room}, {\it reservation}) are usually synonyms. The phrases nonetheless
make immediate sense to familiar listeners, at which point it becomes harder to {\it decompose} the phrase into its constituents.
Perhaps the word ``book'' is there because there used to be physical books that such plans was written down in, and at that
point the reservation was ``booked''. The claim here is not that this is the true etymology, but that the exercise of looking
for such a piecewise explanation is itself unusual for us.

Schr{\"o}dinger \cite{schrodinger1935discussion} famously noted a similar phenomenon in quantum mechanics:

\begin{quote}
When two systems, of which we know the states by their respective representatives, enter into temporary physical interaction due to known
forces between them, and when after a time of mutual influence the systems separate again, then
they can no longer be described in the same way as before, viz. by endowing each
of them with a representative of its own. I would not call that one but rather the
characteristic trait of quantum mechanics, the one that enforces its entire
departure from classical lines of thought. 
\end{quote}

Similar patterns have been noted for centuries --- for example, Aristotle ({\it De Interpretatione}, Ch 2)
argues that in the common phrase ``a fair steed'', the word ``steed'' has an independent meaning, but in the proper name ``Fairsteed'', it does not.
\footnote{The words Aristotle uses are 
\begin{otherlanguage}{greek} Κάλιππσς \end{otherlanguage}
 contrasted with 
\begin{otherlanguage}{greek}
καλός ἵππος\end{otherlanguage},
\begin{otherlanguage}{greek} καλός \end{otherlanguage} as in `calligraphy' and 
\begin{otherlanguage}{greek} ἵππος \end{otherlanguage} as in `hippodrome' or `hippopotamus',
so the correspondence with English `fair steed' is quite direct \cite[p. 116]{cooke1938aristotlecategories}.}
Given the prevalence of composition as a general topic, one might ask whether any overlap between language and quantum
mechanics here is just accidental. One mathematical approach to answering this question has been to explore common
compositional structures using category theory. This has become a fertile direction for research, because common
structures have been identified (in particular, compact closed categories), and moreover, this has shown ways in which
various grammatical and lexical structures can be represented and even combined into tensor networks
that can be implemented as quantum circuits \citep{piedeleu2015open,coecke2020foundations}.

\subsection{Vectors Everywhere}

Vectors and linear algebra have become ubiquitous throughout artificial intelligence, including NLP.
Important cases include the use of feature vectors in statistical machine learning, and the
representation of states and operators in neural networks \citep{geron2019hands}. Linear algebra is also
central to quantum mechanics and quantum computing \citep[Ch 2]{dirac1930quantum,nielsen2002quantum}.

Again one may ask whether these overlaps are significant --- the use of vectors alone might be a shallow
commonality. But many more pieces of quantum mathematics have been found to useful in language-related
tasks, including subspaces and projections for representing conditionals \citep{rijsbergen2004geometry} and
logical negation and disjunction \citep{widdows2004geometry}, quantum probability and the analysis of off-diagonal 
correlations for information retrieval \citep{sordoni_modeling_2013}, entanglement for representing the
combined representation of many pairwise relationships \citep{cohen2012manypaths}. 

\begin{figure}
    \centering
    \includegraphics[width=0.7\linewidth]{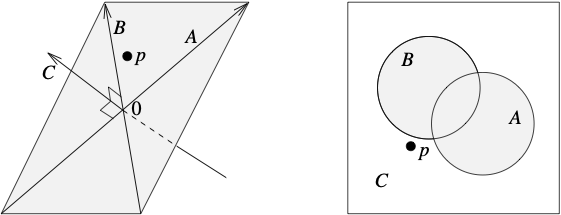}
    \caption{Comparing Quantum (left) with Boolean Logic (right). In quantum logic concepts are modelled
    by subspaces such as lines and planes, and a point can be projected onto any subspace. In classical logic
    concepts can be represented by any set of points, and a point either belongs to this set or it does not.}
    \label{fig:logics}
\end{figure}

While vector spaces have become ubiquitous in AI, some of the opportunities they present have remained comparatively unused.
For example, the lattice of subspaces (lines, planes, and higher-dimensional versions of these) provides a geometric model
for a kind of logic, just as set theory gives a model for Boolean logic (Figure \ref{fig:logics}). This was discovered by
Garrett Birkhoff and John von Neumann in the 1930's
and given the name ``quantum logic'' \cite{birkhoff1936logic}. In terms of operator algebras, the subspaces are in one-to-one
correspondence with the lattice of {\it projectors} onto those subspaces, and projections move points around the space.
This gives quantum logic a {\it non-local} character, which can be exploited for semantic modelling. The negation
operator in quantum logic is projection onto the orthogonal subspace --- for example, in Figure \ref{fig:logics}, 
``{\it p} \textsf{NOT} ({\it A} \textsf{OR} {\it B})'' would be represented as the orthogonal projection of the point $p$ onto 
the line $C$, since $C$ is orthogonal to the plane spanned by $A$ and $B$. This form of vector negation can be used to navigate
between different areas of meaning, rather than just removing unwanted items piecemeal \cite[Ch 7]{widdows2004geometry}. 
Several techniques that have been adapted from earlier quantum literature and applied to problems in AI today.
Many of the successes of
quantum mathematics in artificial intelligence are surveyed by Widdows et al \cite{widdows2021quantum}.

\subsection{The Availability of Quantum Computers}

Far from being considered popular or fashionable, research using quantum mathematical models outside of conventional platforms of quantum physical systems such as isolated atoms have often been met with skepticism and resistance --- perhaps because quantum theory {\it is} too-often overhyped 
and speculative in general science writing, and in some cases because clinging to classical determinism is considered to
be much more scientifically respectable \cite[Ch 1]{orrell2020quantum}. 

These hurdles may be becoming easier to surmount --- for example, major journals have recently published papers 
on quantum approaches in areas including at least economics \citep{orrell2022quantum}, 
psychology \citep{pothos2022quantum}, and AI \citep{widdows2021quantum}. Reasons for this acceptance include
the reality of quantum hardware. Working quantum computers have been built and are running programs every day
--- in a development process spanning decades, concepts like ``entangled pairs'' have gone
from scientific imagination through experimental demonstration to engineering implementation. We may be 
uncomfortable with such quantum concepts, but unlike pioneers in the 1930's, we have evidence that resolves any reasonable
dispute about their reality.

The question for quantum approaches is no longer ``Are they real?'' but ``Are they useful?''. 
In NLP the work is still in its early days, and success must be demonstrated by building useful systems.
Rather than predicting a singular quantum advantage 
for a particular quantum NLP system, this paper explores several examples, with an eye to general capabilities.
The next few sections give examples of how standard 
language processing tasks have been successfully performed on quantum computers. 
Experiments below used the 11-qubit trapped ion quantum computer described by Johri et al. \cite{johri2021nearest},
and where necessary, the larger IonQ Aria machine with a capacity of 32 physical and 20 algorithmic qubits \citep{ionq2022aria}. A crucial point to bear in mind with quantum computing is that the memory capacity
of a system doubles every time a qubit is added, because the number of basis vectors that can be combined
in superposition doubles. While this memory is difficult to 
access in practice and comes with caveats and restrictions, just the notion of exponential memory
growth is a compelling attraction for researchers in data-hungry fields like NLP \citep{widdows2021quantum}.

\begin{figure}
    \centering
    \begin{tabular}{p{7cm}p{5cm}}
    \centering
    \includegraphics[width=2cm,trim={1cm -0.6cm 0.8cm 0},clip]{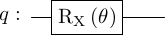} & 
    \includegraphics[width=1.6cm,trim={1cm 0 0.8cm 0},clip]{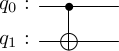} 
    \end{tabular}
    \caption{Basic quantum logic gate diagrams used throughout these examples. A single qubit rotation gate (left)  manipulates superposition of $\ket{0}$ and $\ket{1}$ states for the qubit. The two-qubit CNOT gate (right) entangles two qubits}
    \label{fig:gates}

\end{figure}

Some familiarity with Dirac's bra-ket notation, quantum bits (qubits), gates, and circuits is necessary to fully understand the experiments presented in this paper.
For a thorough review of this area, see Nielsen and Chuang \citep[Ch 4]{nielsen2002quantum}. Alternatively, the more introductory presentation
found in Orrell's textbook \cite[Ch 5]{orrell2020quantum} is enough to understand the new circuits presented below. 
A qubit is a two-level quantum system, whose basis states are normally written as $\ket{0}$ and $\ket{1}$,
and whose pure states are superpositions of the form $\lambda\ket{0} + \mu\ket{1}$, where $\lambda$ and $\mu$
are complex numbers with $\vert\lambda\vert^2 + \vert\mu\vert^2 = 1$. Thus the state of a qubit is represented by a vector in $\mathbb{C}^2$.
When $n$ qubits are assembled, their combined state is represented by the tensor product 
$\mathbb{C}^2 \otimes \mathbb{C}^2 \otimes \ldots \otimes \mathbb{C}^2 \equiv (\mathbb{C}^2)^{\otimes n} 
\cong \mathbb{C}^{(2^n)}$.
The key gate operations are 
single-qubit gates which manipulate the coefficients $\lambda$ and $\mu$ for a single qubit, 
and two-qubit gates, in particular the `controlled-X' or `controlled-NOT' (CNOT) gate that perfoms an X-rotation on the second qubit
of the first qubit is in the $\ket{1}$ state.  Example circuit diagram components for these are shown in Figure \ref{fig:gates}.
Readers for whom these are at least somewhat familiar should be able to understand most of the subsequent examples in this paper.
Readers for whom these are new are encouraged to consult one of the textbooks cited or online 
resources.\footnote{See e.g., \url{https://en.wikipedia.org/wiki/Quantum_logic_gate} and \url{https://qiskit.org/learn/} c.f. \cite{qiskit2021textbook}.}

\section{Word-Based Topic Classification and State Addition}
\label{classification_sec}

Text classification or topic classification is the task of assigning topic labels to texts such as {\sc sports} or {\sc music},
or in the case of ``Which bands performed at the Super Bowl this year?'', perhaps both. It has many uses including
intent classification in dialog systems, assigning topic keywords to documents, routing incoming messages
for customer service, and spam filtering. There are many successful methods, often involving vector representations
and various forms of neural networks and dimensionality reduction \citep{kowsari2019text}.

\subsection{One Qubit Per Word and Topic}
\label{qubit_per_word_sec}

The quantum implementation developed here was deliberately simple: a bag-of-words classifier on a quantum computer.
In this context, `bag-of-words' means that only individual words are used as features for the classifier:
the order of words is not taken into account, or even which combinations of words appear. So for example, a bag-of-words classifier
may predict that a document containing the word {\it horse} is about {\sc farming} or {\sc sports}, which will often
be correct, but is a mistake if it occurs in the phrase ``A horse chestnut is a deciduous tree with palmate leaves.''

The training process for such a classifier is usually some form of supervised machine learning. A collection of texts or documents is sampled,
labels are applied to these texts by human annotators, and these examples are used to train the classifier to deal with
new unseen examples. For a bag-of-words classifier, a simple version of this process is:

\begin{enumerate}
\item For each (document, topic) pair in the training set, split the document text into a list of words.
\item For each word in that list, increment a score recording how often this word was seen with this topic.
\label{question_training_weights}
\end{enumerate}

This process gradually builds up a term-topic matrix (if each document is considered to be a distinct topic,
this is exactly the same as a term-document matrix). The classification stage then proceeds as follows:

\begin{enumerate}
    \item A new document is passed to the classifier. It is split into a list of words.
    \item For each word, look up the scores that word acquired for each topic during training.
    \item Given each topic an overall score accumulated from the sum of the scores for each word.
    \label{question_classifier_sum}
    \item Choose the topic with the highest score as the overall winner.
\end{enumerate}

\begin{figure}
    \centering
    \includegraphics[width=0.4\linewidth]{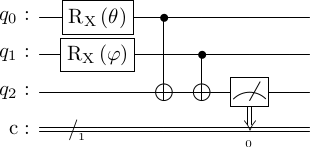}
    \caption{Example Adder Circuit that Combines the Angles $\theta$ and $\varphi$}
    \label{fig:adder_circuit}
\end{figure}

This example was chosen because it raises some basic mathematical implementation questions --- in particular,
item \ref{question_training_weights} in training, ``How do we keep a score for each (word, topic) pair?'' and
item \ref{question_classifier_sum} in classification, ``How do we accumulate an overall score from these individual contributions?''
The first of these problems can be addressed using repeated single-qubit rotations, and the second by connecting
the rotated qubits to a common `sum qubit', as shown in Figure \ref{fig:adder_circuit}.
A detailed analysis of the gate matrices in Figure \ref{fig:adder_circuit} shows that the probability of
measuring a state $\ket{1}$ in the sum qubit $q_2$ is given by $\sin^2(\theta)\cos^2(\varphi) + \cos^2(\theta)\sin^2(\varphi)$.
This is the case when the input rotations are $X$-rotations \cite[\S 1.3.1]{nielsen2002quantum}. Using other fractional rotations
as generators gives combinations with different algebraic properties, investigated more thoroughly in a mathematical paper by Widdows
\citep{widdows2022nonlinear}. Most generally, if the two single-qubit rotations 
in Figure \ref{fig:adder_circuit} are represented by the matrices

\begin{equation*}
\textbf{A} = \begin{bmatrix}  \alpha & \beta \\ -\ov{\beta} & \ov{\alpha} \end{bmatrix} \quad \mathrm{and} \quad 
\textbf{B} = \begin{bmatrix}  \gamma & \delta \\ -\ov{\delta} & \ov{\gamma} \end{bmatrix},
\label{qsum2_eqn}
\end{equation*}
then
\[P(q_2 = \ket{1}) = \vert\alpha\ov{\delta}\vert^2 + \vert\ov{\beta}\gamma\vert^2. \]

The bag-of-words classifier is built by assembling several such qubits and gates. 
During training, each (word, topic) pair
is assigned to a particular qubit, and the (word, topic) weights are accumulated by scanning over the
training set and incrementing the rotation for the corresponding qubit every time a given
word occurs with a given topic. For each topic, an extra qubit is declared to keep the sum for each topic,
so the number of qubits required is $(\mathrm{num\_words + 1})\times \mathrm{num\_topics}$.
(Hence this design only works as a demonstration for very small vocabularies.)
During classification of a new phrase, each recognized
word in the phrase has each of its topic weights connected to the sum qubit for that topic.

\begin{figure}
    \centering
    \includegraphics[width=0.6\linewidth]{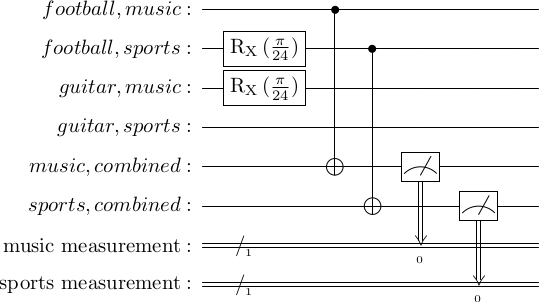}
    \caption{Example Classification Circuit for Two Words and Two Topics}
    \label{fig:classification_circuit}
\end{figure}

As a worked example, consider the toy training corpus:
\begin{center}
\begin{tabular}{lcl}
\\
I play football & $\rightarrow$ & \textsc{sports} \\
I play guitar & $\rightarrow$ & \textsc{music} \\
\\
\end{tabular}
\end{center}
The training process records that the content word {\it football} gives a score to {\sc sports} and {\it guitar} gives a score to {\sc music}
(assuming a preprocessing step that notes that the other words are common to all topics.) 
When classifying the phrase ``Do you want to play football?'', the word {\it football} is recognized and each of the {\it football, topic}
weights are connected to the sum qubit for the corresponding topic. Each of the topic qubits is measured, and the winner is the topic
that measures the most $\ket{1}$ states over a number of shots. A circuit implementing this process is shown in Figure \ref{fig:classification_circuit}.

\subsection{One Qubit Per Embedding Dimension}
\label{embedding_sec}

The design above is clear but wasteful --- a system that requires distinct bits for each word in the vocabulary would be fine in classical
but not yet in quantum computing. In machine learning terms, using a qubit for each word is an example of a `one-hot encoding',
and to avoid redundancy and get more information out of each dimension, more compact distributional vector embeddings are often preferred. The work by Alexander and Widdows \cite{alexander2022quantum} details the method used to classify words from their vector embeddings using a quantum support vector machine (QSVM), and demonstrates subsequent experiments for different QSVM implementations. The steps used in the paper are summarized as follows, and results are explored in further sections.

\begin{figure}
    \centering
    \includegraphics[width=0.6\linewidth]{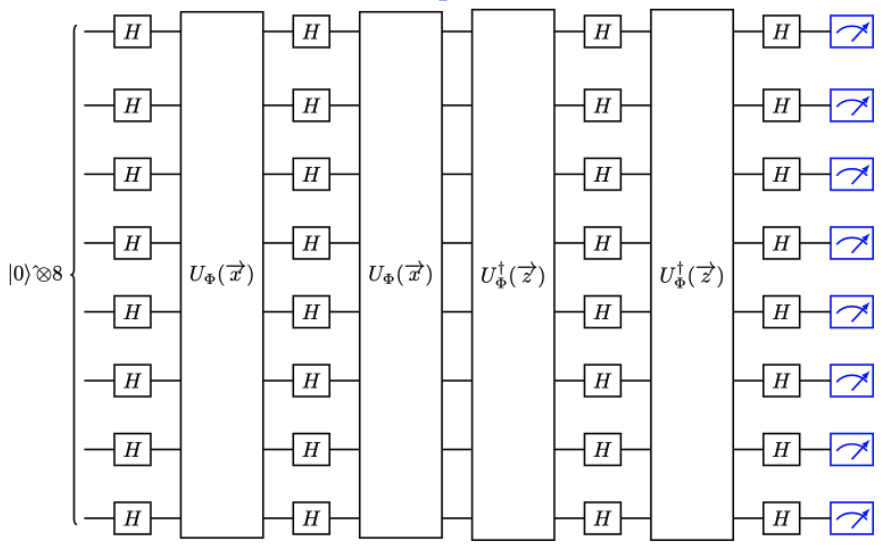}
    \caption{SVM Kernel Circuit Example with 8-Dimensional Inputs \(\vec{x}\) and \(\vec{z}\)}
    \label{svm_circuit}
\end{figure}

First, the Word2Vec technique was used \citep{mikolov2013efficient} to reduce the representation of each word to 8 dimensions.
This part of the process was run classically as a preprocessing step. The generated vectors were then used as input parameters for a feature map quantum circuit. Measurement of the feature map yields a value representing the relationship between two word vectors, which is stored in a kernel matrix \citep{havlicek2019supervise}. This kernel was then used as input for a quantum support vector machine.
For this experiment, the \texttt{ZZFeatureMap} and \texttt{QSVM} implementations from the Qiskit package were used \citep{qiskit2021textbook},
as shown in Figure \ref{svm_circuit}. The \texttt{ZZFeatureMap} applies Hadamard gates on each qubit (denoted by $H$) and a parameterized unitary gate ($U_{\phi}$) in a repetitive cycle; details regarding meaning behind this structure can be found in a preliminary QSVM paper by Havlicek et al \cite{havlicek2019supervise}. The first half of the feature map circuit will use parameters from one word vector, and the second half will take parameters from another word vector.

\subsection{\texorpdfstring{$N$}{N} Qubits for \texorpdfstring{$2^N$}{2\^N} Embedding Dimensions}
\label{dense_sec}

\begin{figure}
    \centering
    \includegraphics[width=0.5\linewidth]{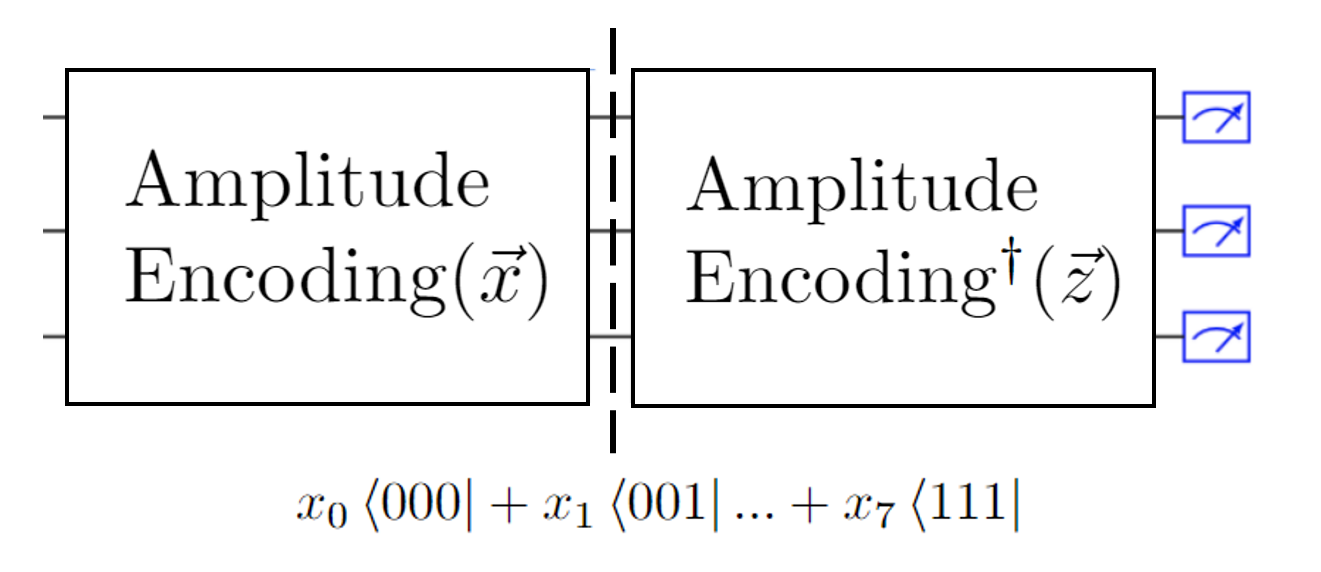}
    \caption{Densely-encoded Kernel Circuit with 8-Dimensional Inputs}
    \label{dense_circuit}
\end{figure}

Building off of an $n$-qubit feature map for $n$-dimensional word vectors, the same QSVM classification process was followed for densely encoded feature maps \citep{alexander2022quantum}. In this case, vector representations were encoded into fewer qubits in the feature map circuit, using \(\log_{2}(n)\)-qubits for $n$-dimensional word vectors. Using such a mapping reduces qubit use without necessarily compromising the complexity of information represented in the word embeddings, as shown in Figure \ref{dense_circuit}.

As observed in the results below, the classical embeddings preprocessing step from Section \ref{embedding_sec} enabled
the quantum circuit to achieve relatively high accuracy with fewer qubits. Further improvements to space using densely encoded feature maps achieved similar results.

\subsection{Data and Experimental Results}

For our initial small-scale experiments, the test and training
datasets from the \texttt{lambeq} package were used \citep{kartsaklis2021lambeq}, which was previously used for the evaluation of the
compositional quantum NLP system of Lorenz et al \cite{lorenz2021qnlp}. This dataset consists of 70 training sentences and 30 development and test sentences, artificially generated to use a small fixed vocabulary and to follow predictable syntactic patterns. 
Lorenz et al.\citet{lorenz2021qnlp} report an accuracy of 83.3\% on the test dataset using a 6-qubit quantum computer, and a similar work by Kartsaklis and Sadrzadeh \cite{kartsaklis2021lambeq}
report achieving 100\% accuracy after several rounds of training on a noisy simulation of a quantum computer.
These are the nearest comparable experiments to those reported here, so the same dataset was deliberately reused.

Both the ``one qubit per word and topic classifier'' of Section \ref{qubit_per_word_sec} was able to reach 100\% accuracy,
and the word embeddings classifiers of 
Section \ref{embedding_sec} and Section \ref{dense_sec} were able to reach 90-100\% accuracy, with comparatively little training and tuning. 

For the bag-of-words classifier of Section \ref{qubit_per_word_sec}, 
the key parameters to tune were the number of words to use and the incremental rotation angle.
The tradeoff here is that too small an incremental rotation can be insignificant compared to statistical noise, 
and too large an incremental rotation can lead to cumulative rotations that go too far and `turn the corner' --- a 
naive version of classical `buffer overflow' in a single qubit. 
The incremental rotation angle was chosen to be $\pi/24$ (as in Figure \ref{fig:classification_circuit}) because this gave
accurate results on the development set. Larger datasets would obviously warrant smaller angles to avoid the overflow issues, and
this parameter tuning could easily be automated for a larger variational learning approach.
The number of words used is partly a tradeoff between space and accuracy. The largest system available to us 
had 20 algorithm qubits \citep{ionq2022aria},
and offline simulations showed that perfectly accurate results could be expected with at least 9 words. To build
a circuit like that of Figure \ref{fig:classification_circuit} with 9 words and 2 topics requires 20 qubits, so 
by a convenient accident, this was just able to run comfortably on available quantum hardware, yielding 100\% accuracy.

For the word embeddings classifier of Section \ref{embedding_sec}, only one qubit per dimension was needed, and 
results were experimentally found to be 90\% accurate with just 8 embedding dimensions, thus requiring only 8 qubits to run. 
When running the densely encoded version of the word embeddings classifier from Section \ref{dense_sec}, 100\% accuracy was achieved using 16 embedding dimensions and only 4 qubits \citep{alexander2022quantum}. This model achieves perfect accuracy for the \texttt{lambeq} set and in the fewest qubits of all the methods covered.

To draw deeper conclusions on scalability, Alexander and Widdows \cite{alexander2022quantum} tested the QSVM classifiers on a more complex dataset of IMDb movie reviews \citep{maas2011imdb}. Actual reviews taken from the database incorporated varied word combinations and colloquial language, 
similar to a realistic use case for quantum NLP. The average number of words in each review is between 228 and 229, and classification experiments were performed in batches of 50 documents: the the number of word tokens involved in each experiment was around 11000 (whereas the total size of 
the \texttt{lambeq} training and test set is 465 words).
The percentage of samples correctly classified peaked when using 4 qubits, where average accuracy was 57\% with the \texttt{ZZFeatureMap} and 62\% using the densely encoded feature map. The QSVM experiments were on par with classical SVM on average, and classified some sample batches with perfect accuracy. This, in contrast with the \texttt{lambeq} results, suggests that good classification merits compatibility of the specific text sample with the feature map or overall method. Such compatibility is not yet predictable and requires further investigation to draw conclusions on the potential of quantum NLP classification on a large scale.

\subsection{Language Caveats and Dependencies}

As well as the simple data size and scaling questions, there are many other language details to be considered, including 
``What does splitting into words mean precisely?'' (tokenization) and ``How do we declare that
two words are the same?'' (canonicalization). These questions were sidestepped for now by using particularly simple
English phrases as the training and test data. However, it is still important to bear in mind that all the approaches
require classical preprocessing of various sorts, including basic whitespace-splitting and assigning tokens to 
qubit indices in the word-based classifier, word vector embedding training for the inputs to the SVM classifier, to the 
need for a full natural language parse tree in the \texttt{lambeq} package.

Rather than claiming that the bags-of-words
classifier is `the best' because it reaches the highest accuracy completely on a quantum hardware, we believe these comparisons
highlight important tradeoffs to consider. With a much more sophisticated linguistic model and related requirements for preprocessing, Lorenz et al.
\cite{lorenz2021qnlp} were able to reach 87\% accuracy with only 6 qubits --- whereas with the same restriction (implemented by 
reducing the number of words considered), the bag-of-words classifier accuracy was reduced to 83\% in simulations.
Having a range of available approaches puts us in a position where tradeoffs between space requirements, system complexity, 
and results accuracy can be considered in the light of what problems users are trying to solve. For example, the work by Alexander and Widdows \cite{alexander2022quantum} investigates solely the effects of decreasing space in the QSVM using a densely encoded feature map. Improved accuracy from 90\% to 100\% in fewer qubits on the \texttt{lambeq} dataset, albeit promising, translated to an unrealistic tradeoff between space and accuracy on the IMDb set. 

For the problem of correctly classifying general text data samples using quantum computation, properly reflecting the complexity of language in quantum representations is challenging. The accuracies from quantum circuits are dependent on the compatibility between the language dataset, type of classes (sentiment, topic etc.), extent of preprocessing and structure of the resulting quantum state. It should be noted that constructing quantum representations for language for any feature-based computation should be purposeful and case-specific, for more effective handling of language dependencies. In this way, the bag-of-words classifier is a more universal approach to handling arbitrary language and classes; it relies only on tokenization compared to abstract feature mappings and word embeddings used by the QSVM.

\section{Bigram Modelling and Quantum Probability}
\label{bigram_sec}

While the ability to combine classifier weights and get good results is important, it is obvious and well-known
that bag-of-words models are a poor approximation for language. Historically, one of the next steps in statistical
language modelling was the modelling language based on short sequences of characters and words, used by mathematician Claude Shannon
\citet[\S3]{shannon1948communication}
to illustrate key behaviors including entropy in the foundation of information theory. Shannon used the
terms `first-order approximation' for a model that samples and generates words based purely on their frequencies, and `second-order approximation' 
for a model that conditions this probability on the preceding word. (For example,  {\it yellow} is likely to come 
before but not after {\it submarine}.) Today these are often called unigram and bigram models, and 
in principle such a model extends to sequences of any length, leading to the long-established field of n-gram language modelling \citep[Ch 6]{manning1999foundations}.
Having built a quantum unigram model, extending this to bigram modelling is a natural next step.

Though still relatively simple, semantic modelling for two-word bigram combinations has real-world applications. For example,
in the hospitality domain, {\it book room}, {\it arrange accommodation}, {\it reserve hotel}, {\it make reservation} all imply
the same intent or goal, even though they do not share the same words and these words are often not synonyms in other contexts
--- for example, {\it book} and {\it make} are not normally synonyms in English. Recognizing that these phrases mean
similar things when combined in this way is this important for intent recognition in customer service dialog systems.

In probabilistic terms, a unigram model can be used for language generation by sampling terms from the distribution
of words, in which case a bigram model works by sampling from the {\it joint} distribution of (ordered) word pairs.
This has an established parallel in quantum probability theory, described in detail by Bradley \cite{bradley2020interface}.
A joint distribution over the vector spaces $V$ and $W$ 
is represented in the tensor product space $V\ot W$, in which case the marginal distributions are represented by
partial traces defined by $\tr_W(v\ot w) = v\, \tr(w)$ and $\tr_V(v\otimes w) = w\, \tr(v).$ 
Bradley \citet{bradley2020interface} uses this approach to represent formal concept lattices in vector spaces.
Formal concept analysis is a theory developed by Georg Cantor \citet{ganter1999formal} in which objects are represented
by which attributes they possess, and concepts arise as closures of sets of objects and attributes that go together.

\subsection{Modelling Steps}

The distributional modelling process outlined by Bradley \citet{bradley2020interface} uses the following steps, 
which are written out fully in the example in Table \ref{trace_model_table}.

\begin{enumerate}
    \item Write down the joint distribution matrix (`Initial Distribution'). In the example in Table \ref{trace_model_table},
    this joint distribution is a subset of the Cartesian product $\mathsf{Colors}\times\mathsf{Fruit}$.
    \item Express this distribution as an element of the tensor product space $\mathsf{Colors}\ot\mathsf{Fruit}$ 
    by flattening the distribution matrix into a long vector. 
    \item Normalize this element to get a distribution state vector $\ket{\psi}$.
    \item Take the outer product $\ket{\psi}\bra{\psi}$ which gives the density matrix of the joint distribution.
    \item Take partial traces of this density matrix. Each partial trace reveals information about the distribution
    of the respective ingredients --- in particular, the diagonal elements represent the individual marginal probabilities,
    and the off-diagonal elements represent information about correlations between items.
\end{enumerate}


\begin{table}[t]
\caption{Steps for modelling the concepts “red apple”, “green apple”, “yellow banana”}
\label{trace_model_table}

\includegraphics[width=\linewidth]{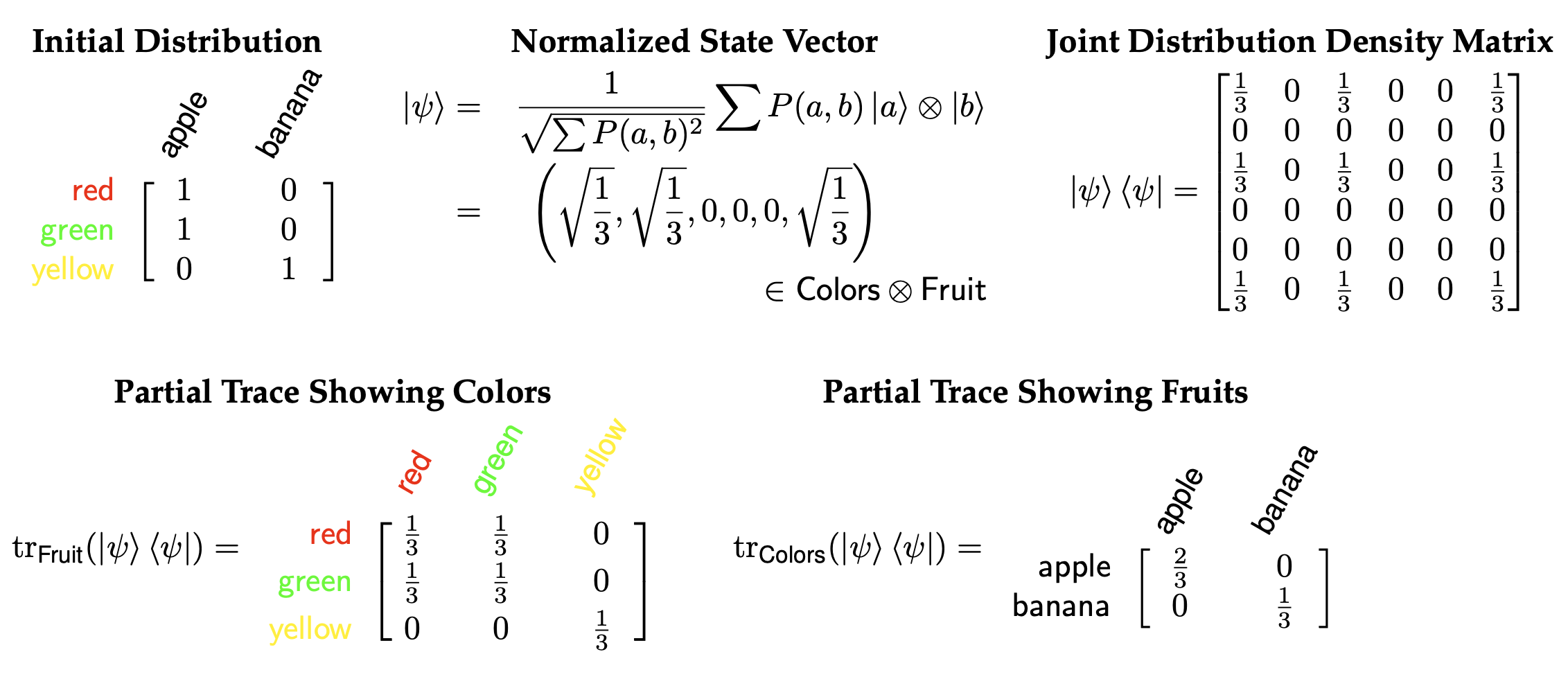}

\end{table}

In the worked example of Table \ref{trace_model_table}, the key thing to note is the off-diagonal $\frac{1}{3}$ terms
that link the {\it red} and {\it green} concepts in the partial trace showing colors. This reflects the fact that back in the original
distribution, these are both colors of {\it apple}. Note that classical marginalization leaves such information behind --- the 
classical marginal distribution would just tell us that {\it red}, {\it green}, and {\it yellow} are equally likely.
This property has been noted by various authors, and was put to work in the work of Sordoni et al. \cite{sordoni_modeling_2013} 
for term-weighting in information retrieval. Those authors recognized that the weight given for a combination of words 
should sometimes be different from the additive weights of the individual words --- taking an example from earlier, 
the weight of the term {\it horse} in the phrase ``I love the flowers on horse chestnut trees'' should be suppressed
because it is unlikely to be relevant for users whose search terms include {\it horse} on its own.

\subsection{Quantum Circuit Approximation for Small Distribution}

The initial quantum implementation goal was to build a joint bigram distribution in a way that can potentially 
be used for language understanding and generation from a probabilistic model. One of the general challenges here is
that preparing a quantum state representing an arbitrary large state vector is difficult --- 
as discussed in Section \ref{dense_sec}, explicitly preparing and comparing vectors 
dimensions requires procedures like the divide-and-conquer method developed by Araujo et al \cite{araujo2021divide}. This worked
reasonably well for embedding 8 feature-dimensions in 3 qubits, but even with a small collection of words,
the corresponding model dimension is much larger, so instead of trying to create the state 
$\braket{\psi}{\psi}$ from Table \ref{trace_model_table} explicitly, we try to learn effective approximations.

For this demonstration, a slightly larger distribution of items and colors was used
including {\it green apple, red apple, yellow banana, ripe banana, red pepper, yellow pepper, black shoes, red dress, blue suit, white shirt}. This has 7 distinct prefixes and 7 distinct suffixes, so to model the joint distribution 
in the manner of Table~\ref{trace_model_table} would require 49 dimensions and $49^2$ for the outer product density matrix.

Instead, in order to fit the distribution into quantum memory and to induce some interference between terms, 
a dense encoding scheme was used.
We use a Quantum Circuit Born Machine (QCBM) model to learn the joint probability distribution. The QCBM encodes the probability of (prefix, suffix) pairs directly as the amplitudes of different states in the superposition. This way, the normalization is satisfied by default. For a joint distribution with $n$ prefixes and $m$ suffixes, we would need $nm$ classical bits just to model `present or absent'. For the QCBM model,
we use $\lceil\log_2(n)\rceil + \lceil\log_2(m)\rceil$ qubits, and an encoding convention where each prefix and suffix is given a binary index,
and the cooccurrence weight is stored as the amplitude of the state given by concatenating these two indices.  
Each of the adjectives and nouns was assigned a binary 
index corresponding to a tensor product state, for example $\textit{blue} = \ket{101}$ and $\textit{shoes} =\ket{100}$.
These are concatenated to form the bigram index, $\textit{blue shoes} = \ket{101100}$.

\begin{figure}
    \centering
    \includegraphics[width=\linewidth]{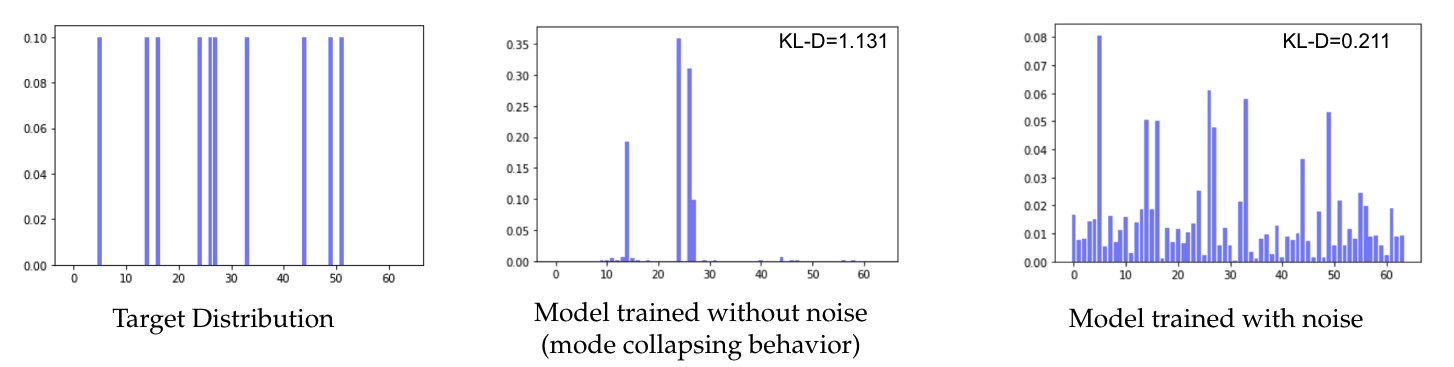}
    \caption{Attempts to fit the target distribution, with and without introducing noise. The goodness-of-fit is measured by Kullback-Liebler (KL) divergence, 
    $D_\text{KL}(P \parallel Q) = \sum_{x\in\mathcal{X}} P(x) \log\left(\frac{P(x)}{Q(x)}\right)$, where $P$ is the learned distribution and $Q$ is the target distribution.}
    \label{bigram_model_fits}
\end{figure}

The QCBM uses parametric quantum circuits to generate different superpositions. The training challenge is then to find the proper set of parameters that lead to the generation of the specific superposition that represents the target joint probability distribution. For this, a
Simultaneous Perturbation Stochastic Approximation (SPSA) optimizer  \citep{spall1998overview} was implemented which optimizes the parametric quantum circuit, minimizing the difference between the generated distribution and the target distribution using the Kullback–Leibler (KL) divergence as the cost function. Sample results are shown in Figure \ref{bigram_model_fits}.
The initial results were disappointing (middle column) --- while finding some peaks in the distribution, the optimizer
entirely missed others, leading to a poor KL-divergence score of $1.131$. Our hypothesis was that the target distribution
was too discrete or `spiky' to be fitted effectively in this way. 
Further experiments were carried out that {\it added} noise to the distribution and tried to fit the noisy version.
Perhaps surprisingly, this worked considerably better --- the optimizer was able to fit most of the spikes in the distribution, at the cost
of giving some weight to other elements, achieving a much better KL-divergence of $0.211$. 

This result is analogous to the use of smoothing in $n$-gram models. It is often observed
that language produces new words and phrases that have not been seen before: this is what is meant when language is said to be
`generative'. To accommodate this, language models are adapted to give non-zero probabilities even to phrases that have never been encountered in training,
and there are several traditional ways of doing this \cite[Ch 6]{manning1999foundations}. 
In a simple demonstration, after training, the model was used to generate a few new phrases by random sampling. Results included some plausible
example like {\it black suit}, and occasionally misfits such as {\it ripe shirt}.

\subsection{Quantum Circuit Approximation for Larger Cue-Target Distributions}

This section investigates how these quantum circuits and results behaved at a larger scale with more realistic data.
For this, a larger collection of pairs was needed, for which the University of South Florida Free Association Norms 
dataset was used \citep{nelson2004university}. This dataset collects many results of experiments where subjects were
given a prompt-word, such as {\it add}, and responded with a word they consider associated, 
such as {\it subtract} or {\it calculate}. This dataset has been used to compare different models for analyzing word-meaning
and retrieval \citep{bruza2009there}, and to propose word-pairs for subsequent word-similarity ranking experiments \citep{chandrasekaran2021evolution}. After combining the input files, the word-graph used in our experiments had
1703 distinct cue words, 6359 distinct target words, and 24526 links from a cue to a target. Each link was weighted by the 
forward-strength of the link as computed by Nelson et al \citet{nelson2004university}.

\begin{figure}[ht]
    \centering
    \includegraphics[width=0.8\linewidth]{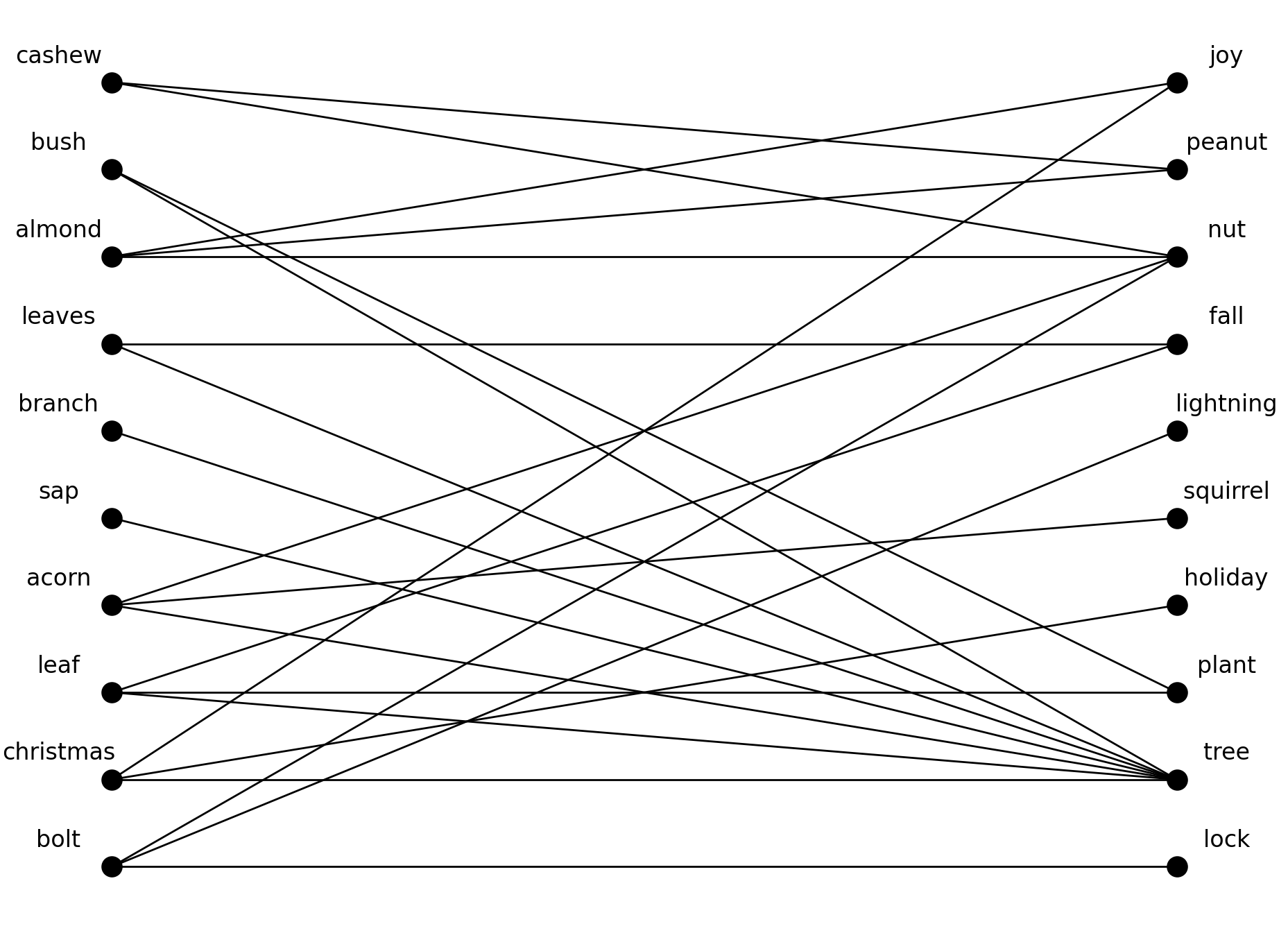}
    \caption{Bipartite cue-target graph grown around the seed-word {\it acorn}}
    \label{fig:acorn_graph}
\end{figure}

Given $n=1703$ and $m=6359$, it would need $\lceil\log_2(n)\rceil + \lceil\log_2(m)\rceil = 24$ qubits to attempt to 
model the whole graph distribution, which even with advances in recent months is beyond the current size of reliable
quantum computers \citep{ionq2022aq}. This introduces the requirement of sampling from the larger graph to get representative subgraphs. 
A natural problem here is that taking a random sample of nodes and retaining the edges between them does not preserve representative graph properties 
--- for example, if $p$ nodes are randomly sampled from a graph with $N$ nodes, the chance of a node $n$ being selected is $\frac{p}{N}$, 
whereas the chance of an edge $(n_1, n_2)$ being selected is roughly $\frac{p}{N^2}$ because both $n_1$ and $n_2$ need to be selected.
Graph sampling challenges and example solutions are surveyed by Leskovec and Faloutsos \citet{leskovec2006sampling}: in these terms, the graph sampling challenge
would be described as a kind of scale-down sampling, and the method used is related to the random walk family. More precisely, 
a subgraph is grown by starting with a small set of seed nodes, scoring their neighbors according to the sum of edge-weights 
connecting them with the seed nodes, and adding batches of the $k$ neighbors with the best scores until the desired number of nodes is reached.
In our bipartite graph this process alternates between cues and targets, so the same number of each is sampled. An example subgraph 
with 10 cues and 10 targets starting with the seed word {\it acorn} is shown in Figure \ref{fig:acorn_graph}.

\begin{table}[t]
\renewcommand\arraystretch{1.3}
\scriptsize
    \centering
    \begin{tabular}{|l|ccc|ccc|ccc|}
    \hline
    Nodes in Graph   & & 16 &&& 32 & \\
    & Prec. & Rec. & F1 &Prec. & Rec. & F1 \\
    \hline
SPSA          & \bf{0.062} & \bf{0.047} & \bf{0.053} & \bf{0.042} & \bf{0.029} & \bf{0.034}  \\
SPSA Smoothed & 0.041 & 0.033 & 0.036 & 0.025 & 0.019 & 0.022 \\
Shared Links  & 0.054 & 0.045 & 0.049 & 0.032 & 0.025 & 0.028 \\
Random        & 0.046 & 0.034 & 0.039 & 0.040 & 0.029 & 0.033 \\

    \hline
    \end{tabular}

\vspace{0.15in}

    \centering
    \begin{tabular}{|l|ccc|ccc|ccc|}
    \hline
    Nodes in Graph   & & 64 &&& 128 & \\
    & Prec. & Rec. & F1 &Prec. & Rec. & F1 \\
    \hline
SPSA          & 0.022 & 0.016 & 0.019 & 0.011 & 0.008 & 0.009  \\
SPSA Smoothed & 0.015 & 0.011 & 0.013 & 0.008 & 0.006 & 0.006  \\
Shared Links  & \bf{0.051} & \bf{0.039} & \bf{0.044} & \bf{0.056} & \bf{0.041} & \bf{0.048}  \\
Random        & 0.020 & 0.015 & 0.017 & 0.011 & 0.008 & 0.009  \\

    \hline
    \end{tabular}

\vspace{0.15in}

    \caption{Precision, recall, and F1-score of link recovery experiments for graphs up to 128 nodes. Each 
    figure is the average of 100 experiments with graph grown around a randomly chosen seed word.}
    \label{tab:link_generation_eval}
\end{table}

This technique was used to create a 100 sample subgraphs containing each of 16, 32, 64, 128 modes, by choosing 
a random cue-word (sampling uniformly from all cues) and growing a subgraph around that word. The pattern of 
being able to approximate target distributions more effectively with some smoothing was confirmed ---
for example, the average KL-divergence between the learned and the target distributions with 16 nodes was 
1.504 without smoothing and 0.454 with smoothing.

A further experiment was performed to see how these learned distributions perform at generating new links.
This was simulated by removing 20\% of the links in the sampled subgraphs in a random train-test split, training
the models on the 80\% of training links, taking the pairs with the highest probabilities in the learned distribution that 
were not links in the training graph, and comparing these with the 20\% of held-out test links. 
The results were evaluated using the standard scores of precision (number of correct links divided by number of 
proposed links), recall (number of correct links divided by number of test links that could be found), and F1-score
(the geometric mean of precision and recall) \cite[Ch 3]{geron2019hands}. The results of this experiment are 
shown in Table \ref{tab:link_generation_eval}. In spite of the the better KL-divergence scores, the SPSA
results with smoothing are consistently less good than those without, and are sometimes worse than the random
baseline scores. For small graphs (16 and 32 nodes), the SPSA method performs better than randomly, and better than
an alternative Shared Links approach, which uses a similar link-scoring technique to that used for selecting the
sample subgraphs to rank the most promising available new links. 
However, for the larger graphs (64 and 128 nodes), the Shared Links method pulls ahead.

These results are somewhat disappointing, though representative of some of the challenges quantum NLP
is likely to face. Mathematically exciting models may lead to promising implementations, from which we can
learn a lot about distributional patterns and propose improvements. However, classical computers can
still process much more data quickly so in many cases a relatively simple classical baseline
algorithm can still produce better results.

\section{Ambiguity Resolution in Composition}
\label{ambiguity_sec}

The ability to choose appropriate interpretations for ambiguous words is a hallmark of human language capability.
Much of the time, we guess meanings quickly and correctly, but various experiments have also shown that in some 
situations, several meanings are not only available but cognitively activated, causing and confusion and delay
in human language responses \cite[Ch 19]{aitchison2002words}.
\footnote{Sometimes ambiguity resolution is so effortless that we underestimate the prevalence of ambiguity itself, 
which takes many forms. For example, English has considerable syntactic ambiguity --- one large sample found that 
57\% of the 400 most common English verbs are also nouns \citep{widdows2003taxonomies}, but we rarely notice this.
As noted by \cite{baroni2014frege}, ``Nearly all
papers in formal semantics start from the assumption that all the words
in the examples under study have been neatly disambiguated before
being combined. But this is not an innocent or easy assumption.''}

Word-sense discrimination and disambiguation were among the early successes of word vector models during the 1990's, 
built using dimension-reduction of term-document or term-term cooccurrence matrices \citep{schutze1998automatic}. This was 
followed by the use of orthogonal projection as a logical operator for isolating particular senses of ambiguous 
words, confirming that a semantic vector for an ambiguous word can be analyzed as a sum of 
vectors for difference senses, which can sometimes be recovered from the geometry of the
vector space \cite[Ch 7]{widdows2004geometry}. 

The ambition of combining distributional word vectors with compositional operations that model grammatical structures
led to successful experiments with matrices as operators --- for example, where nouns are modelled as vectors, adjectives
can be modelled more effectively as matrices acting upon the noun vectors \cite{baroni2010nouns}. Putting
such insights together enabled comprehensive systems for distributional compositional semantics to be devised 
\citep{coecke2010distributional,baroni2014frege}, leading to many successful experiments 
(for a more thorough summary see Widdows et al. \cite{widdows2021quantum}). 

As noted by authors including Baroni and Zamparelli \cite{baroni2010nouns}, the use of matrices to operate on vectors supports disambiguation
as different components of meaning are selected in the compositional process. For example, we might learn that {\it red}
means something closer to {\it orange} in `red hair', something closer to {\it purple} in `red wine', and something closer
to {\it urgent} in `red alert'.

\begin{table}
\caption{A toy domain model with column vectors for nouns and matrices for transitive verbs}
\label{nouns_verbs_tables}
\vspace{0.1in}
\begin{tabular}{ccc}
\begin{tabular}{|r|ccccc|}
\hline
&
\begin{turn}{70} \sf{Borneo} \end{turn} &
\begin{turn}{70} \sf{Java} \end{turn} &
\begin{turn}{70} \sf{C++} \end{turn} &
\begin{turn}{70} \sf{Trip} \end{turn} &
\begin{turn}{70} \sf{Juggling} \end{turn} \\
\hline
\sf{Place} & 1 & 1 & 0 & 0 & 0 \\
\sf{Event} & 0 & 0 & 0 & 1 & 0 \\  
\sf{Tech} & 0 & 1 & 1 & 0 & 0 \\  
\sf{Skill} & 0 & 0 & 0 & 0 & 1 \\  
\hline
\end{tabular}

&

$
\sf{visit} = 
\begin{bmatrix} 
0 & 0 & 0 & 0 \\ 
1 & 0 & 0 & 0 \\ 
0 & 0 & 0 & 0 \\ 
0 & 0 & 0 & 0 \\
\end{bmatrix}
$

&

$
\sf{learn} = 
\begin{bmatrix} 
0 & 0 & 0 & 0 \\ 
0 & 0 & 0 & 0 \\ 
0 & 0 & 0 & 0 \\ 
0 & 0 & 1 & 0 \\ 
\end{bmatrix}
$

\end{tabular}

\end{table}

To demonstrate such a process on quantum computer, we designed the example domain shown in Table \ref{nouns_verbs_tables}.
There are only four dimensions, representing the broad areas {\it place}, {\it event}, {\it technology}, {\it skill}, and 5 nouns --- one belonging to
each dimension, and the noun {\it Java} which could refer to a place or a technology. Consider now the transitive verbs {\it visit} and {\it learn}.
When we say ``I visited Borneo'', this indicates that an event happened --- so the verb {\it visit} takes places as input and creates events as output.
When we say ``I learned C++'', this indicates that a skill was acquired --- so the verb {\it learned} in this case takes technologies as input
and creates skills as output. It is easy to write down matrices that perform these operations (Table \ref{nouns_verbs_tables}, right hand side).

Consider now the action of these matrices on the ambiguous word {\it Java}. Both the {\it place} and the {\it tech} dimensions are activated
with this input, but the verb {\it visit} simply ignores the {\it tech} part of the input, whereas the verb {\it learn} ignores the {\it place} part.
In the process of composition, the meanings that naturally combine with one another are preserved, and those that do not are discarded.

\begin{figure}
\centering
\renewcommand{\arraystretch}{1.5}
\setlength{\tabcolsep}{0.2in}
\begin{tabular}{ccc}
Circuit for {\it Java} &
Circuit for {\it visit} &
Circuit for {\it visit Java} \\
    \includegraphics[valign=t,width=0.95in,trim={0 -0.1cm 0 0.25cm}]{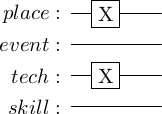} & 
    \includegraphics[valign=t,width=0.9in]{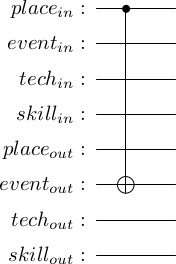} & 
    \includegraphics[valign=t,width=1.56in,trim={0 -0.2cm 0 0.25cm}]{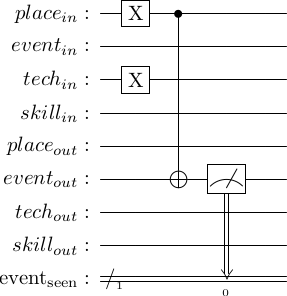} \\
    
\end{tabular}

    \caption{Toy circuit examples showing how the phrase {\it visit Java} selects the {\it place} meaning}
    \label{visit_java_circuit}
\end{figure}

This process was then modelled using the quantum circuits in Figure \ref{visit_java_circuit}. Qubits corresponding to the dimensions for {\it Java} are activated by applying the $X$ gate, and the verb {\it visit} is implemented
by connecting the input {\it place} qubit to the output {\it event} qubit. The quantum circuit is actually more complicated mathematically than
the small matrices of Table \ref{nouns_verbs_tables} --- for example, the state of the four input qubits would be represented in 
$(\mathbb{C}^2)^{\ot 4}\cong \mathbb{C}^{16}$. These interactions would have to be considered when modelling a larger domain, if only
because the approach in Figure \ref{visit_java_circuit} would soon run out of qubits. Nonetheless, an experiment with these circuits
on the 11-qubit IonQ machine gave the expected results --- measuring the output bits for the {\it visit Java} or the corresponding {\it learn Java}
circuits shows that the former activates the {\it event} output and the latter activates the {\it skill} output. Ambiguity is an example
of a problem that is crucial to language understanding, to which the mathematical tools of vectors and matrices have been successfully applied,
and the circuits of Figure \ref{visit_java_circuit} provide an example of how such processes can be implemented on quantum computers.

\section{Related Work}
\label{background_sec}

These experiments and results come in the midst of much progress in quantum machine learning and AI.
Until recently, most work in quantum algorithms was theoretical in how these algorithms
could not be run on quantum hardware. This has changed rapidly, and in the 2020's,
new results from quantum computers in artificial intelligence are being published almost every month, 
with applications including probabilistic reasoning \citep{borujeni2021quantum}
financial modeling \citep{johri2021nearest}, and image classification \citep{wang2021experimental,johri2022generative}. 
Similar questions arise across these areas, partly because the availability of 
some of the key quantum properties like indeterminacy, interference, and entanglement  
pose the question ``How can these mathematical properties help to model a given situation?''

The most directly comparable research to that presented here is the compositional quantum approach
released in the \texttt{lambeq} package \citep{meichanetzidis2020qnlp,kartsaklis2021lambeq}. This is the ongoing culmination of
a research program outlined by Coecke et al. \citet{coecke2010distributional}, which is often referred to as the DisCoCat
(`distributional compositional categorical') model. This combines the use of distributional semantic vectors
with syntactic parsing, and partly follows the tradition of logical semantics in the sense that the semantic
representation is derived from first finding the syntactic or grammatical structure of a sentence. 

Such an approach to meaning is common in theoretical linguistics, which often considers
sentences as fundamental units of language and require grammaticality. The relationship between
categorial grammar and compositional semantics especially is a key part of Montague grammar \citep{partee1976montague}.
Such direct predecessors in NLP today are sometimes obscured by the tremendous success of 
machine learning methods since the early 2000's: as the field has 
coalesced around a few particularly successful techniques, other topics that would once have been
included in a standard introduction to language and meaning have become less well-known.
A renewed interest in logical semantics encouraged by its use in DisCoCat could bring insights
independently of its potential applications in quantum NLP. 

At the same time, precedents in logical semantics also bring reservations --- starkly, for example,
in 1976 philosopher David Lewis wrote,
{\it ``It is apparent that categorial grammars of this sort are not reasonable grammars for natural language''}
\citep{lewis1976general}. Lewis' concerns were partly about the inability of phrase-structure grammars to
account for freer word-orders, which motivates the addition of a transformational component. This helps to 
adapt the machinery to grammatical structures in more languages, but maintains the assumption that systems
are working with grammatical sentences to begin with. 
While this is still appropriate in many formal domains, the pervasive availability this century
of mobile electronic devices has led to a deluge of informal text which also (and frequently!)
demands our attention.

The challenge of inferring grammatical structure from informal text could itself become an opportunity
for QNLP, since early studies suggest that language parsing can be done efficiently on quantum computers \cite{wiebe2019quantum}. This raises the possibility that grammatical structure could interact with
the semantic compositional structure in a model such as DisCoCat as quantum information, rather than 
as a fixed piece of classical information from a preprocessing step, which could make the system
more efficient and effective when working with incomplete or indeterminate parses.

An approach
based more on information retrieval \citet{rijsbergen2004geometry,widdows2004geometry}
starts bottom-up with individual words instead of grammatical sentences, and moves from there to pairs of words \citep{sordoni_modeling_2013}.
The examples in this paper have more obviously drawn on this research, partly because we believe it 
will be applicable to more language situations in general.
Another family of 
techniques that combines syntax and semantics has been to train deep neural networks
and distill both semantic and grammatical information together from these \citep{palangi2017grammatically}. 
The use of tensor product decomposition networks has particularly interesting overlaps with models
used in quantum chemistry \citep{mccoy2020tensor}. 

However, the most advanced language systems
in widespread use currently are neural networks such as GPT-3 that do not use explicit grammatical preprocessing
at all,
and these models are so fluent that critiques of them focus on semantic and ethical failures instead 
of anything related to grammar and syntax \citep{floridi2020gpt}.
Human grammatical fluency is still scientifically interesting, but the lack of it is no longer a weak-link 
in language processing systems: language generation systems can easily produce  
``massive amounts of very plausible but often untrue texts'' \citep{sobieszek2022playing}. For this reason we
expect aligning semantics and factual accuracy to be a more pressing engineering challenge 
than aligning semantics and syntax.

Given the huge range of language challenges in computing, and pace of development
in both language processing and quantum computing, we expect 
that the next few years will witness many more contributions and combinations.

\section{Conclusions and Next Steps}
\label{conclusion_sec}

The experiments and demonstrations presented in this paper show that NLP tasks can be performed on
quantum computers, which is important in itself --- quantum information processing is being demonstrated on 
quantum hardware, not just in theory. The practical applications are so far small-scale, and it is still challenging to
find consumer-facing NLP requirement that would be better served today by quantum than classical hardware.
And as demonstrated in some of our experiments, a more elegant mathematical method may produce worse results
than a simple baseline model that can train on more realistic data.

The main dependency here is hardware: because quantum memory capacity grows exponentially with the number of qubits,
10 qubits can represent a kilobyte, whereas 30 qubits can represent a gigabyte (roughly speaking --- this compares
the number of variables, not their expressivity or access patterns, and classical memory is still much more reliable). 
This threshold is likely to be crossed in the next two years, which will enable us to
explore larger datasets, and to look for useful intermediate-scale applications. 

The scaling challenge is particularly important in NLP, because one of the consistent findings in this paper
is that promising results with very small curated datasets do not always translate into good performance on larger,
more varied and realistic datasets. This challenge is analyzed in detail by Alexander and Widdows \citet{alexander2022quantum}, 
where it shown that the that enough data can be processed to get results that are reliably above baseline
in a statistical sense, but still much less than is regularly used in classical NLP systems.

Though much of the focus in NLP research in recent years has been on larger and larger models, many 
commercial systems involve much smaller curated datasets, such as the intents used by dialog systems
or ticket classifications used by customer service tracking systems. With a dedicated focus on building useful products, 
it should become possible for quantum NLP to help with practical medium-term goals, and this can help show the
way to more general intelligent systems. 
\backmatter

\section*{Data Avaliability}

The datasets that were analyzed in the current study may be found at the following public domain resources:
\begin{itemize}

    \item {\it Lambeq}: CQCL Github repository at \url{https://github.com/CQCL/lambeq}
    \item {\it IMDb}: \url{https://ai.stanford.edu/~amaas/data/sentiment/}
    \item {\it University of South Florida Free Association Norms}: Available through the supplementary material at \url{https://doi.org/10.3758/BF03195588}
\end{itemize}

\noindent
Conflict of Interest: The authors declare that they have no conflict of interest.

\bibliography{ionq}

\end{document}